\documentclass[conference]{IEEEtran}
\IEEEoverridecommandlockouts
\usepackage{gensymb}
\usepackage{cite}
\usepackage{url}
\usepackage{amsmath,amssymb,amsfonts}
\usepackage{algorithmic}
\usepackage{graphicx}
\usepackage{textcomp}
\usepackage{xcolor}
\usepackage{float}
\def\BibTeX{{\rm B\kern-.05em{\sc i\kern-.025em b}\kern-.08em
    T\kern-.1667em\lower.7ex\hbox{E}\kern-.125emX}}
\usepackage{geometry}
\usepackage[tight,footnotesize]{subfigure}
\usepackage{verbatim}

\begin{document}

\newgeometry{top=1in,bottom=0.75in,right=0.75in,left=0.75in}
\title{GenGrid: A Generalised Distributed Experimental Environmental Grid for Swarm Robotics }



\author{\IEEEauthorblockN{Pranav Kedia}
\IEEEauthorblockA{IIIT-Bangalore\\
pranav.kedia@iiitb.org}
\and
\IEEEauthorblockN{Madhav Rao}
\IEEEauthorblockA{IIIT-Bangalore \\
mr@iiitb.ac.in}
}

\maketitle

\begin{abstract}

GenGrid is a novel comprehensive open-source, distributed platform intended for conducting extensive swarm robotic experiments. The modular platform is designed to run swarm robotics experiments that are compatible with different types of mobile robots ranging from Colias, Kilobot, and E-puck~\cite{colias1,colias2,kilobot,monarobot}. 
The platform offers programmable control over the experimental setup and its parameters and acts as a tool to collect swarm robot data, including localization, sensory feedback, messaging, and interaction. GenGrid is designed as a modular grid of attachable computing nodes that offers bidirectional communication between the robotic agent and grid nodes and within grids. The paper describes the hardware and software architecture design of the GenGrid system. Further, it discusses some common experimental studies covering multi-robot and swarm robotics to showcase the platform's use. GenGrid of 25 homogeneous cells with identical sensing and communication characteristics with a footprint of 37.5~cm $\times$ 37.5~cm, exhibits multiple capabilities with minimal resources. The open-source hardware platform is handy for running swarm experiments, including robot hopping based on multiple gradients, collective transport, shepherding, continuous pheromone deposition, and subsequent evaporation. The low-cost, modular, and open-source platform is significant in the swarm robotics research community, which is currently driven by commercial platforms that allow minimal modifications.

\end{abstract}


\section{Introduction}
The study of swarms and collective robots is advancing towards smaller systems, and there is a dearth of tools needed to study the behavioral states of an individual or a group in motion. In addition to exploring new algorithms in a simulated environment while collecting enough data and validating the same on a physical robot space. Many swarm robotic platforms~\cite{colias1,colias2,kilobot,monarobot} studied in the past, offer infrastructure consisting of agent-to-agent communication. However, there is a need for smaller and simpler systems that makes experiments like~\cite{kiloassem} possible, where 1024 Kilobots~\cite{kilobot} were programmed to aggregate into predefined shapes.
However, the simplicity of these small robots with minimal resources and a low-cost solution is often a limiting factor in these experiments. Environment-based communication through Stigmergy-enabled systems is one of the alternate solutions to overcome the limitations of current swarm robots. Swarm robotics provide unique benefits over centralized approaches that include robustness due to massive redundancy via multiple agents, and scalability due to distributed load~\cite{swarm1, swarm2}. Stigmergic communication in swarm robotics is derived from herd animals' implicit cue messaging technique, based on sensing specific characteristics and following the group~\cite{Hind2013, HeikoBook}.  

The asynchronous communication in Stigmergy that exploits the environment is considered highly effective, especially for small-sized animals. The same is recently extended to the physical form of robotic systems. Stigmergy in ant-based pheromone trails is considered highly suitable for achieving effective communication in small-sized swarms. The platform to satisfy and validate micro-level communication between agents and macro-level objective representing a task by the swarm group is a challenge, considering minimal resources to mimic the limited functionality between agents yet achieving the broader objective. The challenge is to design an independent swarm platform to demonstrate macro-level objective and robot design compatible with the platform for performing and validating swarm experiments, with an understanding of micro-level information and visually rendering global outcomes~\cite{236,192,239,Reina2015}.
Incorporating global and local features in implementing a swarm design platform needs particular attention and cannot be addressed by purely inheriting distributed systems solutions. Besides, implementing and demonstrating the same on a hardware system in conjunction with a platform of minimal resources is a sound challenge, which is the focus of this paper. The proposed implementation separates the macro and micro level attention towards platform and robot space individually and cohesively demonstrates the swarm level activities. The individual-specific implementation for a robot to plan its activities based on micro-level observations requires an optimal and minimal resource-driven robot design. Similarly, for demonstrating the overall goal, a robust platform with visual representation is appreciated. Although most of the swarm activities are well characterized in graphical user interface simulators~\cite{218, 142}, yet the robot mimicking swarm activities needs a physical implementation before the final deployment. The physical implementation, especially in a smaller or indoor environment, is highly suitable for characterizing swarm behavior. Several robotic platforms are designed for swarm experiments in the past, including Khepera~\cite{H11}, Khepera III~\cite{H12}, e-puck robot~\cite{H13},
Alice~\cite{H14}, Jasmine~\cite{H15}, I-swarm~\cite{H16}, S-Bot~\cite{H17}, 
Kobot~\cite{H18}, SwarmBot~\cite{H19}, Kilobot~\cite{kilobot}, and  Colias~$\phi$~\cite{colias1,colias2}. 
The Stigmergy based bio-inspired robotics is a highly efficient swarm group, considering minimal resources, optimal activity per individual agents, and achieving the given task~\cite{khaliqstig}. 

%
The stigmergy approach was implemented on physical robots~\cite{khaliqstig} by building wavefront navigation maps, which were then permanently stored on an RFID floor, and maintaining distance to the closest obstacle in the RFID table which is computed by a second stigmergy algorithm. The combined use of wavefront and distance to obstacle map allowed any robot to perform goal-directed navigation while maintaining a safe distance from obstacles without using any of the self-localization approach and minimal computational load. 
The computationally efficient and low-cost robot design is susceptible to the RFID module installed on the platform. Hence, hardware-established robot and platform design that is less sensitive to the individual sensor modules is preferred. Another work of~\cite{cosphi,colcosphi} presents an open-source artificial pheromone system that is robust, accurate, and uses readily available components, including an LCD screen and low-cost USB cam. This system can simulate many pheromones and changes parameters on the fly to investigate agents' dynamic interactions and overall outcomes. However, the additional camera support for robot interactions reflects overutilization of resources and inherently does not truly represent pheromone behavior characterized by small animals. The Pheromone-based communication has shown to be used in swarm robotics both in simulation-based~\cite{cos4} and real-robot based studies~\cite{swarm1}. In some real-robot based studies, alcoholic chemical substances were used to simulate pheromones~\cite{cos6,cos7,cos8,cos9}. However, a chemical disposing agent deprives the swarm design of the control of diffusion and evaporation of a leaving trail, leading to a limitation in mimicking pheromone behavior. Additionally, the chemicals used are flammable and are not safe to use during unsupervised experiments. Additionally, few artificial chemical pheromones 
that are expensive are studied in the past; however, realizing these approaches
to build swarm platform is highly infeasible ~\cite{cosphi,colcosphi}. Literature shows us that the simulation of pheromones was carried out by employing RFID tags \cite{cos10,khaliqstig}, audio sources \cite{cos12,cos13}, and light sources \cite{cos14,cos15,cos16,cos17}. These methods are very flexible, easy to use, and readily available compared to their chemical counterparts. However, RFID tags' finite-size limits the resolution of the simulated pheromone trails~\cite{cosphi, colcosphi}. The audio signals, too, are inflexible for programming evaporation and diffusion of pheromone trail, in addition to being sensitive to outside noise. Light shows flexible characteristics in emitting different intensities from low-cost components~\cite{cos17}, and was employed effectively to study pheromone-based communication with micro autonomous robots using projected light~\cite{cos18, cos17}. The light-based pheromone communication offered high-resolution trails, controllable diffusion and evaporation processes, and the ability to achieve aggregated or suppressed tasks. Hence light acting as a pheromone is used in the proposed GenGrid platform to enable robot to robot and robot to platform bi-directional communication channel. The paper proposes GenGrid, a comprehensive open-source reconfigurable, modular, scalable, and low-cost platform mimicking pheromone communication via light source for swarm robotics experiments, with minimal resources. 
Two robots (of the same type) inspired by the current swarm robot design~\cite{kilobot}  were built to validate various experiments on the platform. Each robotic agent exercises a magnetic effect on the cell to imprint the existence in the grid. The novel GenGrid platform allows software reconfiguration to perceive illuminated cells as artificial obstacles, or foraging actuators, or implicit interactions.





\section{System design}

\begin{figure}[ht!]
\centering
    \subfigure[]{ \includegraphics[scale = 0.45]{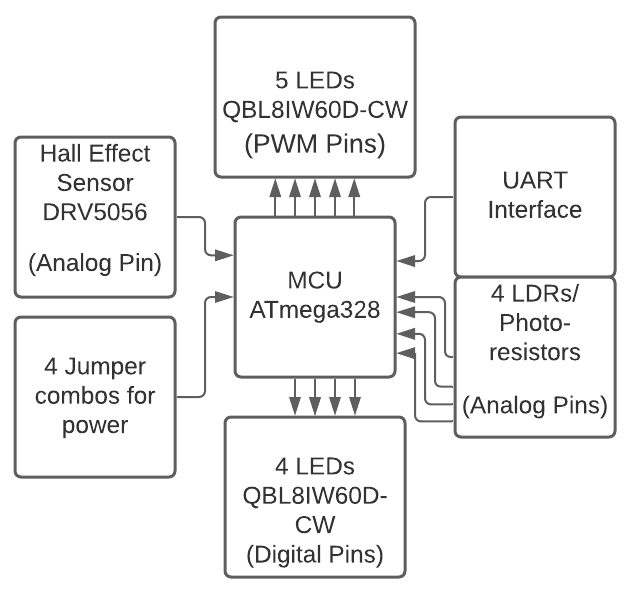}}
    \subfigure[]{\includegraphics[scale = 0.40]{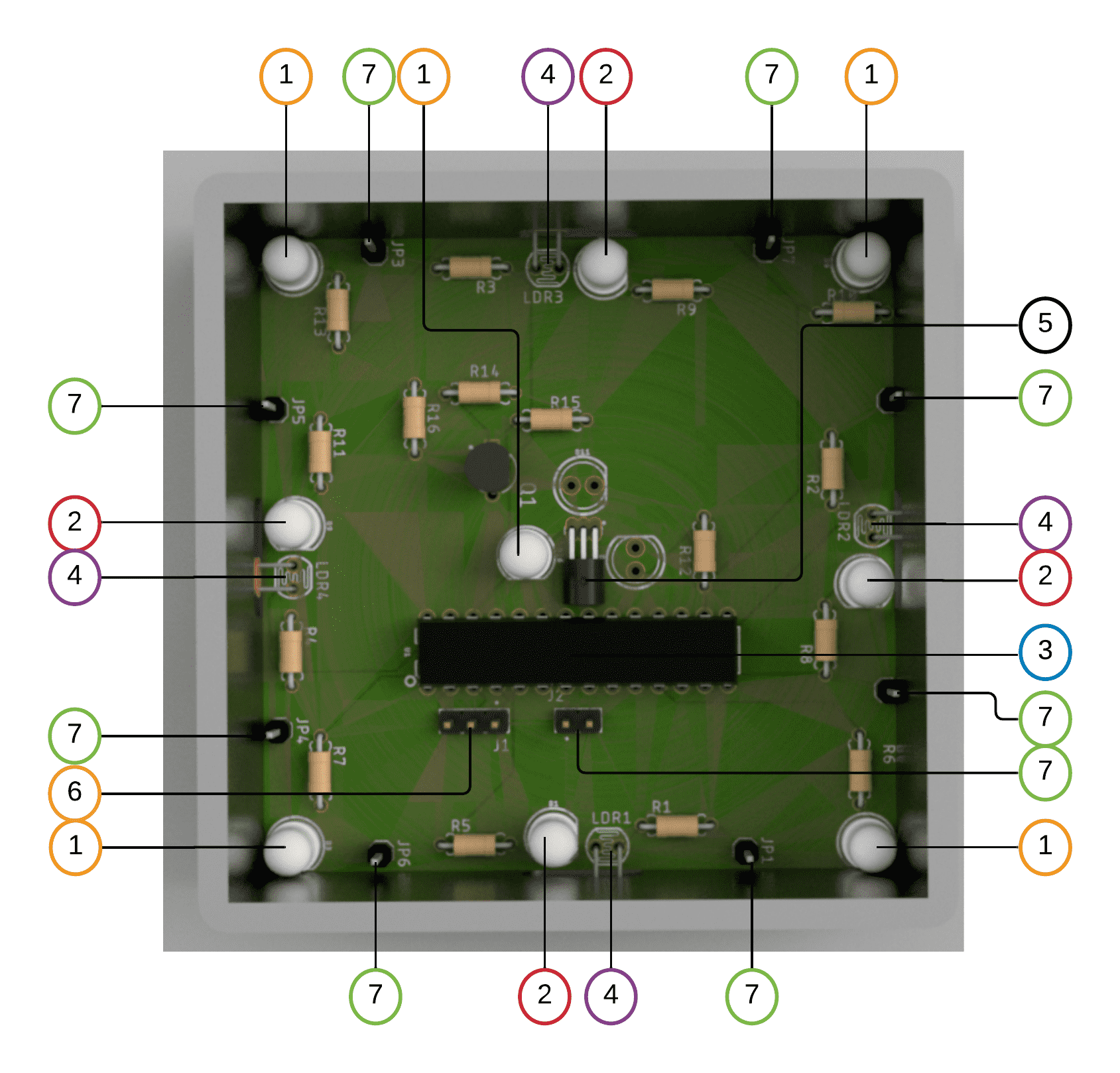}}
    \subfigure[]{\includegraphics[scale = 0.40]{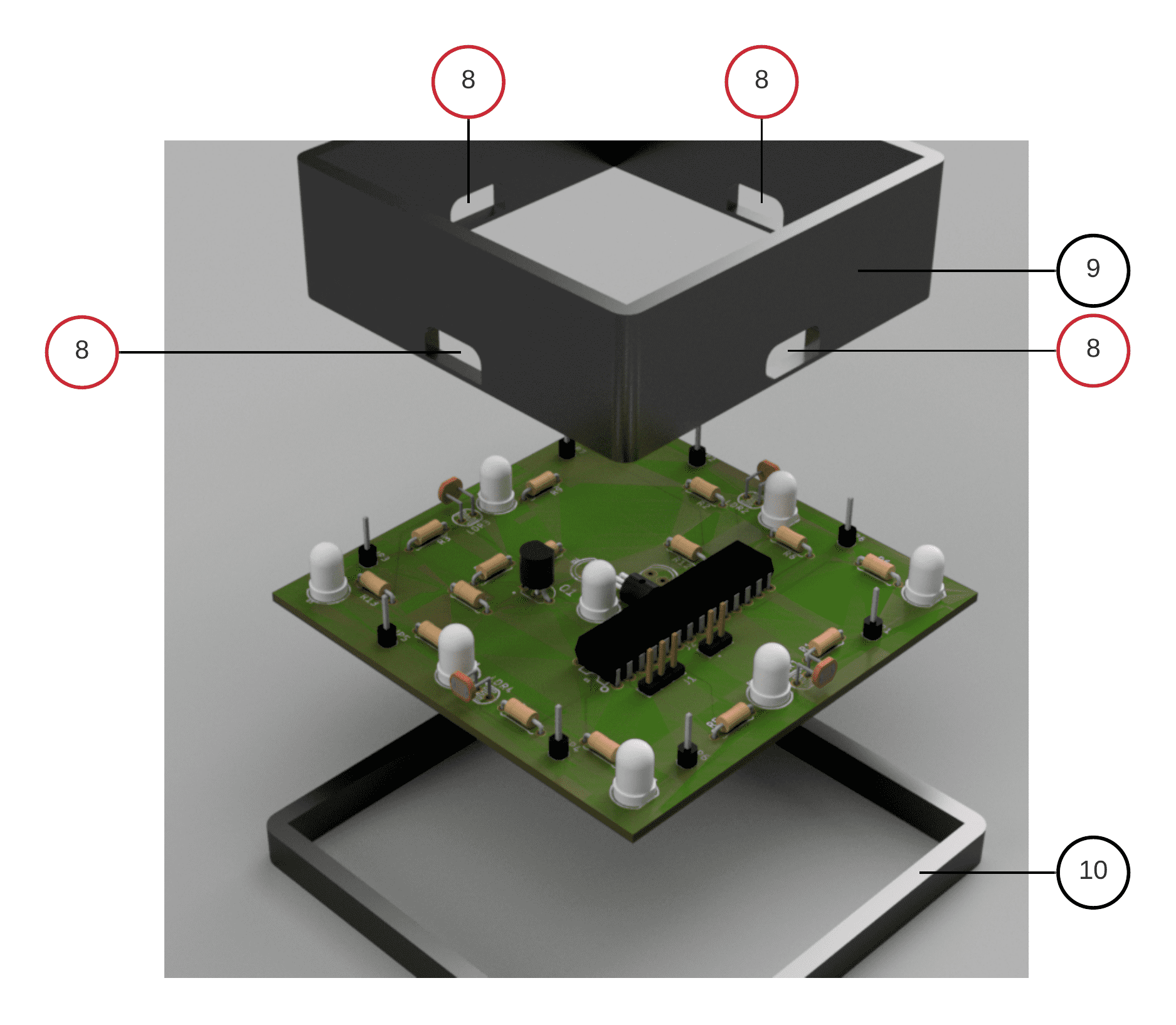}}
    \caption{Architecture design of GenGrid module showing
    (a)block level design with all components of a GenGrid cell, (b)2D design of GenGrid cell, and (c) exploded 3D render of digital GenGrid cell.}
    \label{fig:gengrid}
\end{figure}

\begin{figure}[htp]
    \begin{center} 
    \includegraphics[scale = 0.05]{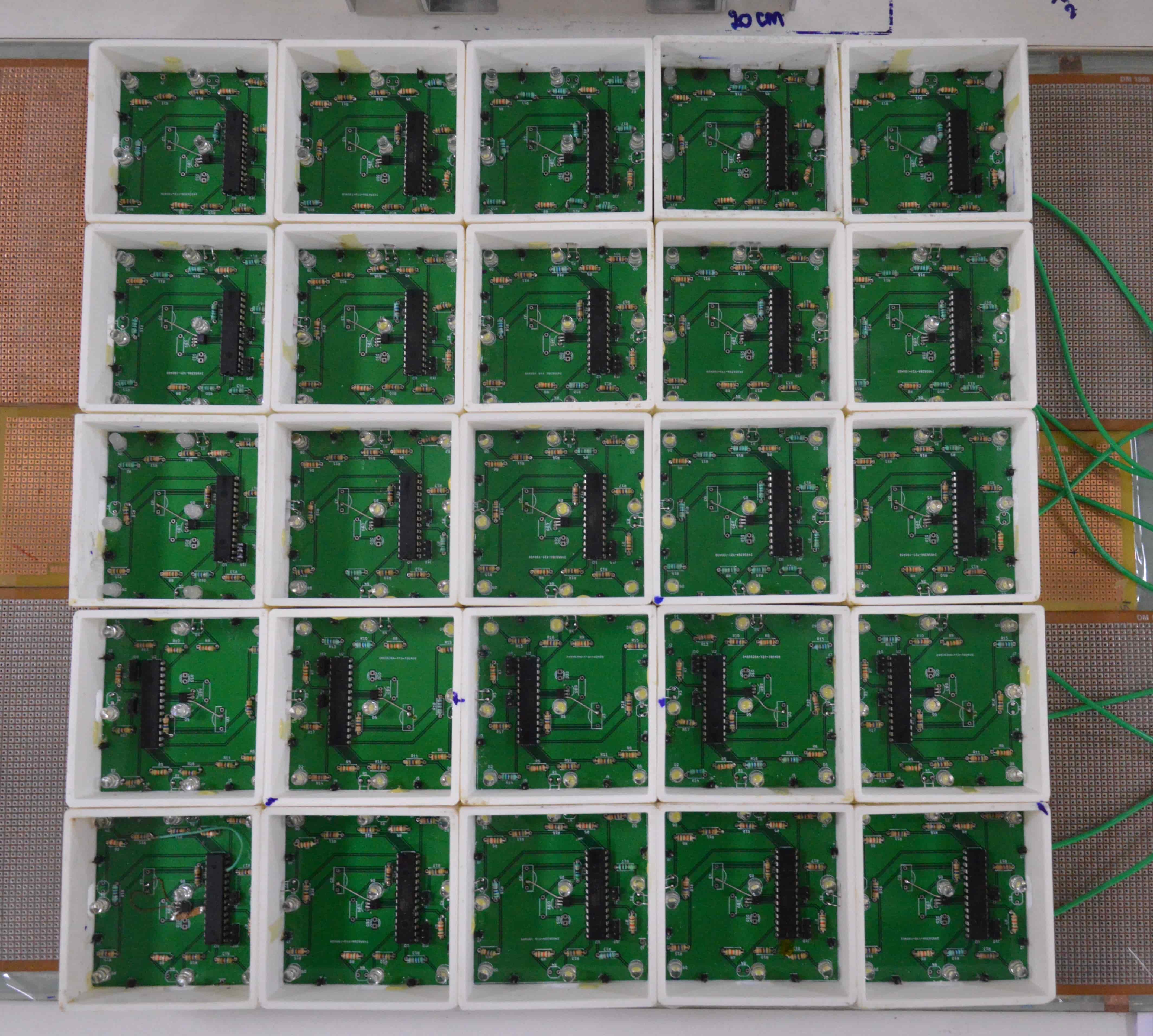}
    \caption{Screenshot of the developed GenGrid platform.}
    \label{fig:array}
    \end{center}
\end{figure}

GenGrid, a comprehensive open-source reconfigurable, modular, and scalable platform of 25~modules with a footprint of 375~mm $\times$ 375~mm, is designed and developed for swarm robotics experiments.
Each GenGrid cell, as shown in Figure~\ref{fig:gengrid}~(a), is equipped with sensing capability, as well as
communicating requisite signals to the other robotic agents running in the platform. An analog magnetic hall effect sensor is
included in the cell design to sense the presence of a swarm robot on the grid. Five LEDs driven by Pulse width modulated (PWM) digital pins are wired in the grid module to enable showcasing the passing swarm robot trail.
The PWM-controlled LEDs are positioned around the four corners of a square module and one at the center.
Besides, four digital LEDs are designed at the sides of the module to enable inherent von Neumann neighborhood messaging within the GenGrid platform.
The integrated photoresistors/LDRs allow reading the behavior of the neighborhood whenever configured. 
Thus the combination of a magnetic sensor and LEDs enables the communication between robot to the grid cell and the swarm world, whereas 
the LDRs and LEDs in the grid cell allow communication between the GenGrid cells.
The UART interface enables easy programming of the individual cells to furnish the required features of sensing and lighting for an experiment. 
The prototype architecture with individual components is shown in Figure~\ref{fig:gengrid}~(a).
Figure~\ref{fig:gengrid}~(b,c) shows a top view and 3D exploded digital render of the GenGrid module with labels numbering from 1 to 10 to satisfy the following objectives:
\begin{enumerate}
    \item The PWM-controlled LEDs used as light actuators for implicit communication with the robot.
    \item Digital LEDs used as light actuators for explicitly programmed communication with von Neumann structured neighborhood cells.
    \item MCU ATmega328P is a microcontroller used for programming the individual cells.
    \item The LDRs as receivers to read the von Neumann neighborhood cells state.
    \item The Hall effect sensor DRV5056 used as an implicit receiver for messages from the robot on top.
    \item Serial/UART interface for the programmer. This can be replaced with a CAN bus network to allow for simultaneous programming and data collection, as shown by ~\cite{kilogridjournal}.
    \item Power delivery configured with five pairs of Vin and GND connectors.These act as power delivery conduits from one cell to its neighbors allowing us to connect large no. of grid cells like LEGO bricks. 
    \item Communication cutouts for allowing light-based communication between the cells. 
    \item Top 3D printed shell ~\cite{S1}.
    \item Bottom 3D printed shell ~\cite{S1}.
\end{enumerate}
The developed platform consisting of 25 modules is shown in Figure~\ref{fig:array}. A thin glass layer is laid on top to close the platform and allows swarm robot movement on the top.
The transparent glass top allows minimum loss of magnetic field from the 
robot, and minimum optical signal dilution.
The GenGrid design with a glass top and 3D printed enclosures on the sides makes the system stable and robust for swarm robotic experiments.
The minimal and low-cost resources in the form of LEDs for light-based communication and magnetic sensor for sensing robot traces fit the requirement of designing a novel swarm robot platform.
The overall cost of an individual GenGrid cell is around 8 USD, effectively costing 200 USD for a 25 cell GenGrid platform.

\begin{figure}[ht!]
    \begin{center}
    \subfigure[]{\includegraphics[scale = 0.026]{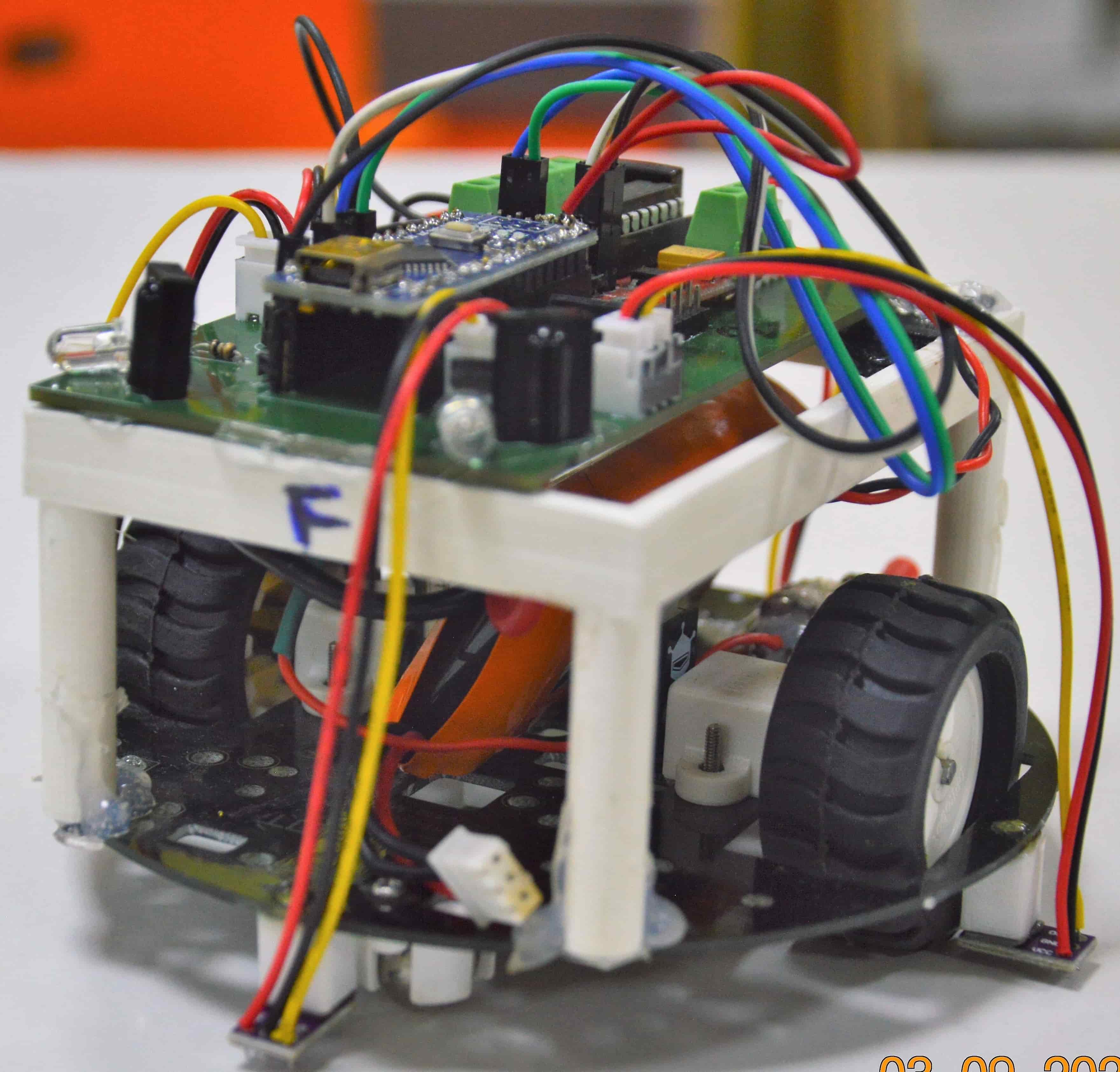}}
    \subfigure[]{\includegraphics[scale = 0.52]{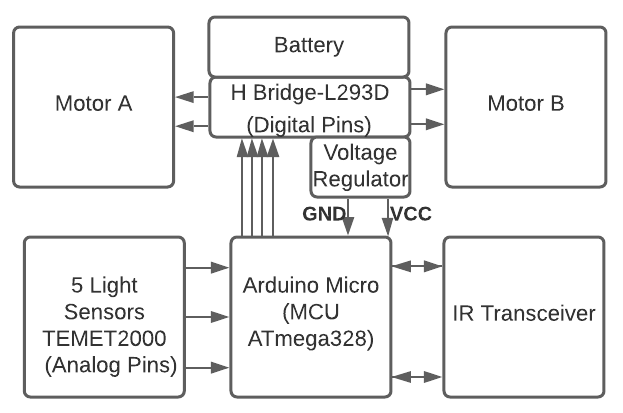}}
    \caption{Robot design showing (a) screenshot of the developed swarm 
    robot, and (b) block level drawing of the robot showing all components.}
    \label{fig:robot}
     \end{center}
\end{figure}

An Arduino-based robotic platform inspired by currently popular swarm robotics platforms~\cite{kilobot} is developed to showcase experiments using the novel and low-cost GenGrid platform. The robot's image is shown in Figure~\ref{fig:robot}~{a} and a block diagram showing all the components and their interfaces is mentioned in Figure~\ref{fig:robot}~{b}. The robot is capable of both implicit and explicit communication with other swarm members and the GenGrid platform. The robot is equipped with a permanent magnet at the bottom to communicate its presence to the GenGrid module. A disc-shaped neodymium magnet
of 20~mm diameter and 10~mm thick is fixed at the bottom of the robot. Use of Gengrid requires the robot to be bigger than the individual cell. Hence, the robot is built on a 122mm diameter chasis which enabled a clever placement of 5 light sensors on the current cell and the neighboring cells.
The robot is designed with sensing light and actuating magnetic fields to have a symbiotic relationship with the developed GenGrid platform, mimicking the pheromones trailing behavior of ants.
The onboard five light sensors sense the light actuation of varying intensity from the GenGrid module and its four neighborhoods.
The differential drive robot is built on a two-wheeled DC motor and chassis, which gives two degrees of freedom (DoF), which is appropriate for movement in the GenGrid platform. The onboard IR transceivers can be used to communicate explicitly with other swarm robots in the future.
Thus robot to robot, direct communication is possible due to IR transceivers, 
along with a robot to GenGrid bi-directional communication is
feasible using light sensor and magnet attached.
A two LiPo cell battery of 600 mAh is used to power the robot.
The overall cost for the robot design is around 20 USD, which is close to other popular platforms. 
The GenGrid platform is designed such that any swarm robot with magnet holding capability at the bottom is compatible with the GenGrid platform.

\section{Experiments and Results}

\begin{figure}[htp]
    \subfigure[]{\includegraphics[scale = 0.30]{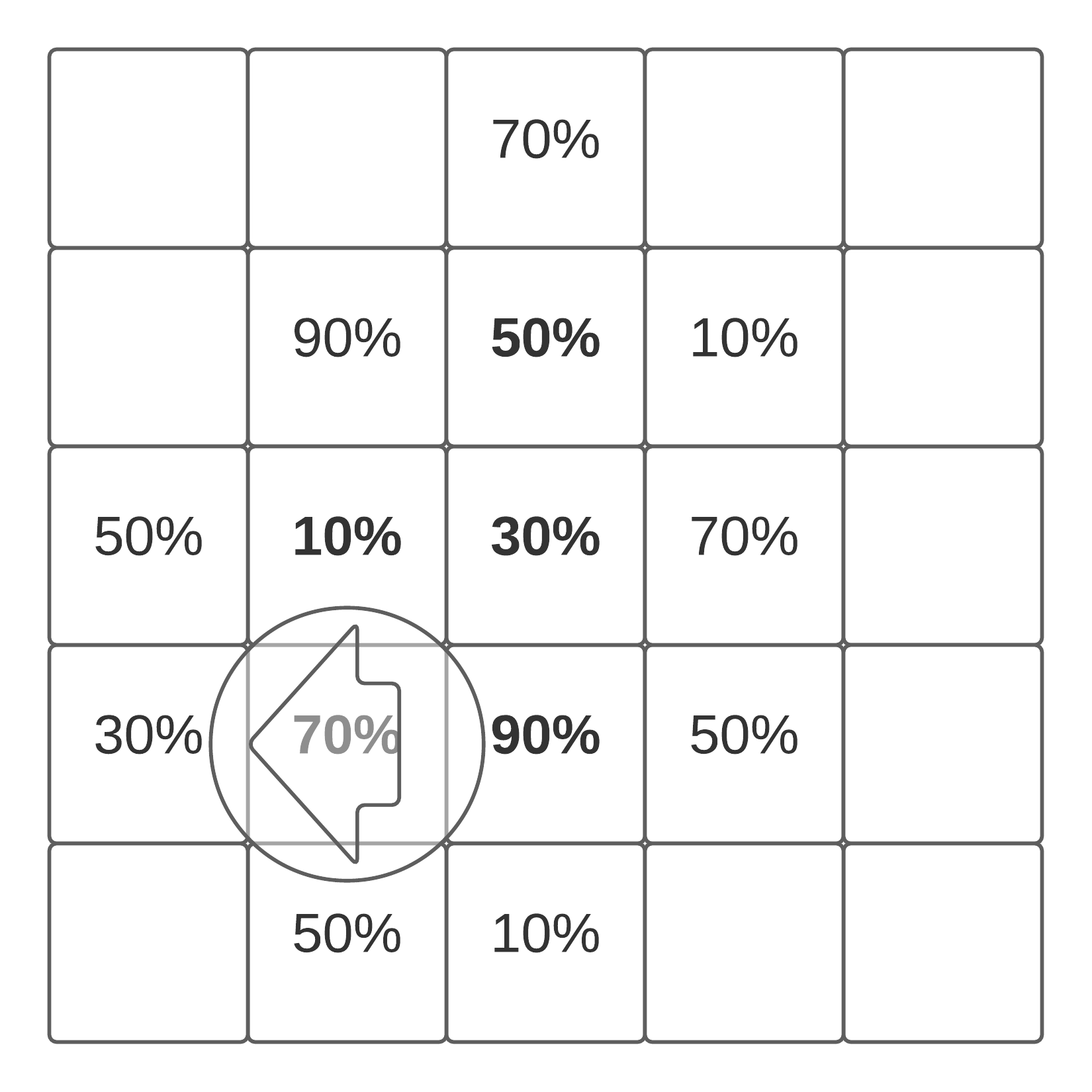}}
    \subfigure[]{\includegraphics[scale = 0.029]{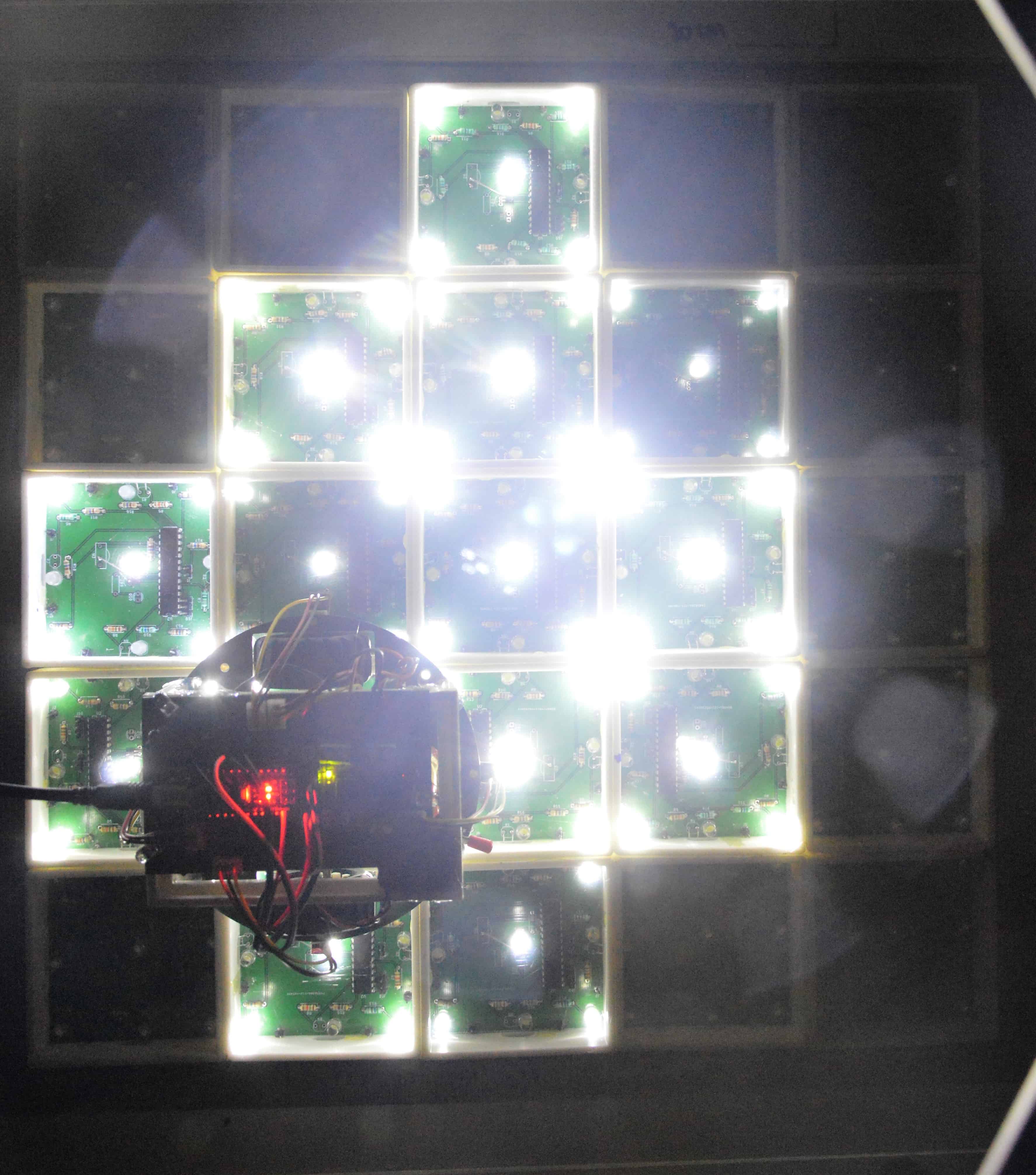}}
    \subfigure[]{\includegraphics[scale = 0.25]{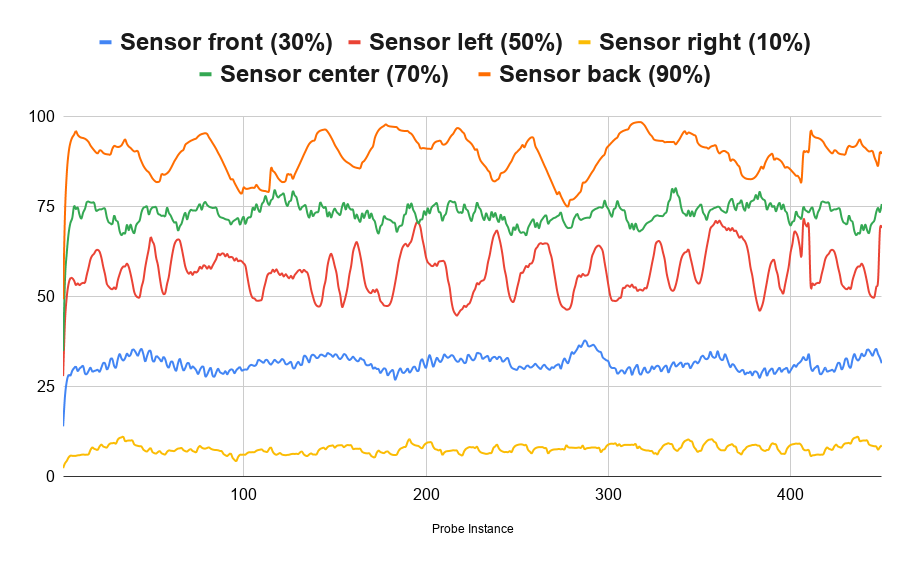}}
    \caption{(a) Map specifying the intensity of light at the center of
individual cells with robot positioned at 70\% intensity, (b) image showing the screenshot of the actual GenGrid configured system with robot placed on top,
and (c) Light sensor readings from the positioned robot in the configured
GenGrid platform.}

    \label{fig:exp1}
\end{figure}

The robot sensing the light intensity of GenGrid cells forms the critical aspect
of conducting swarm experiments. Hence the programmable GenGrid platform is configured with fixed light intensity values, and robot readings using light sensors were measured at three different positions in the platform. One such case is shown in the Figure~\ref{fig:exp1}. 
The robot with five light sensors positioned on the front, back, left, right, and center were used to measure the illumination received by each of them. The light sensors at the sides of the robot measure the intensity of light falling from the neighborhood, and at the center measures the illumination
of the present cell.
The acquired sensor values are shown in the (a) part of Figure~\ref{fig:exp1}. 

The acquired sensors values at each of the different positions of the robot,
validates the illumination programmed in the GenGrid platform and the robot's capability to measure relative light intensity data.

\begin{figure}[htp]
    \subfigure[]{\includegraphics[scale = 0.273]{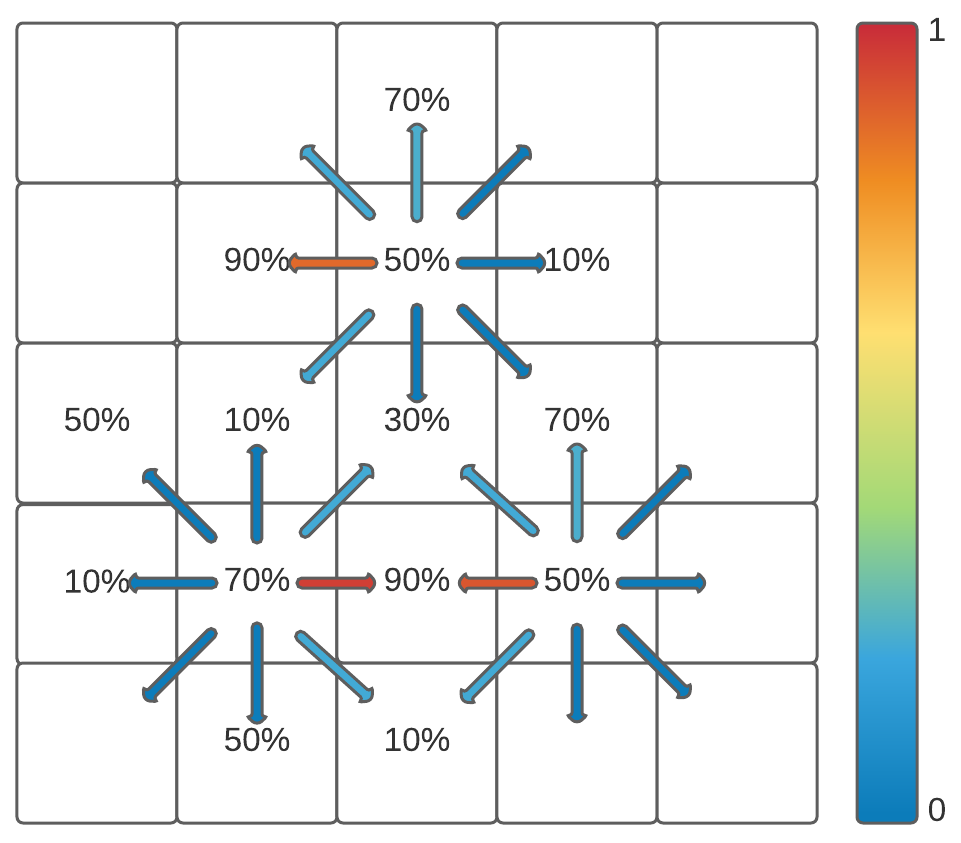} \label{fig:singlehop}}
    \subfigure[]{\includegraphics[scale = 0.273]{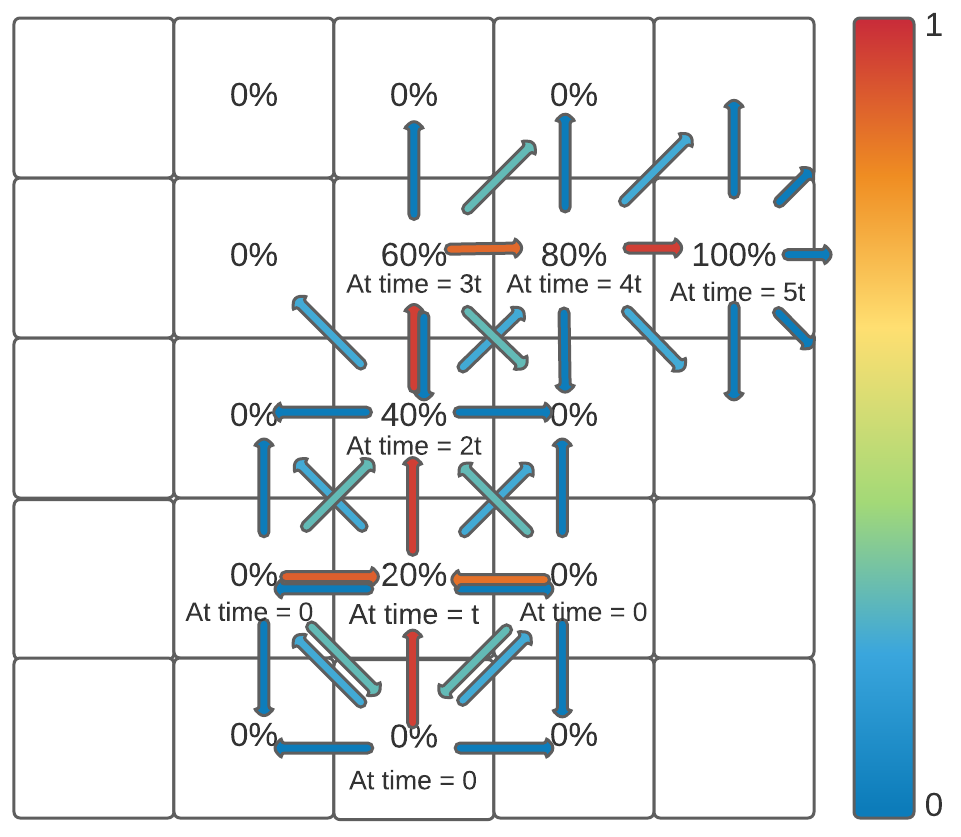} \label{fig:multihop2d}} 
    
    \caption{(a) Image showing the probability of robot position in a GenGrid map
    while performing single hop based on positive light gradient, (b) Image showing the probability of robot position in a 
    GenGrid map while following 2D path based on positive light gradient.}
\end{figure}

Different experiments were carried out to demonstrate various possibilities of realizing swarm-based pheromone behavior using the GenGrid platform.
For validating the single hop experiment on GenGrid, the cells were programmed with different light intensities, and the robot's movement at different positions is evaluated.
The robot was programmed to move along the maximum positive gradient in the grid and the von Neumann neighborhood. 
The experiments of a robot performing single-hop were repeated by placing the robot in different grids 20 times each, and probability 
distribution for the robot to hop to the next cell for three
different starting positions, where light intensity was maintained at 70\%, 50\%, and 50\% each, were performed. The experiment results are depicted in Figure~\ref{fig:singlehop}, with arrows representing the hopping direction, and the color of the arrows represents the probability of the robot moving in that direction to the next cell. A color bar specifying the probability gradient is also included in the figure. Light intensity in percentage is mentioned within the cell.
In all three experiments, the probability of the robot moving to the expected and highly illuminated cell around 90\%.
The failures represented by other arrow directions showcased significantly less probability and are attributed to open-loop robotic design with the robot deviating away from the target cell.
The probability of the robot moving to the highest intensity cell was 90\%, 
a bit lower than expected.
However, considering all the 3 cases where the robot was positioned, the
the robot had to undergo rotational motion and then move to the next cell. The open-loop 
rotational motion induces deviation and hence concedes 
10\% error in reaching to
the next expected cell position.
The single-hop robot experiment is considered as a first step towards 
robot movement in an interactive environmental platform.

\begin{figure}[htp]
    \includegraphics[scale = 0.20]{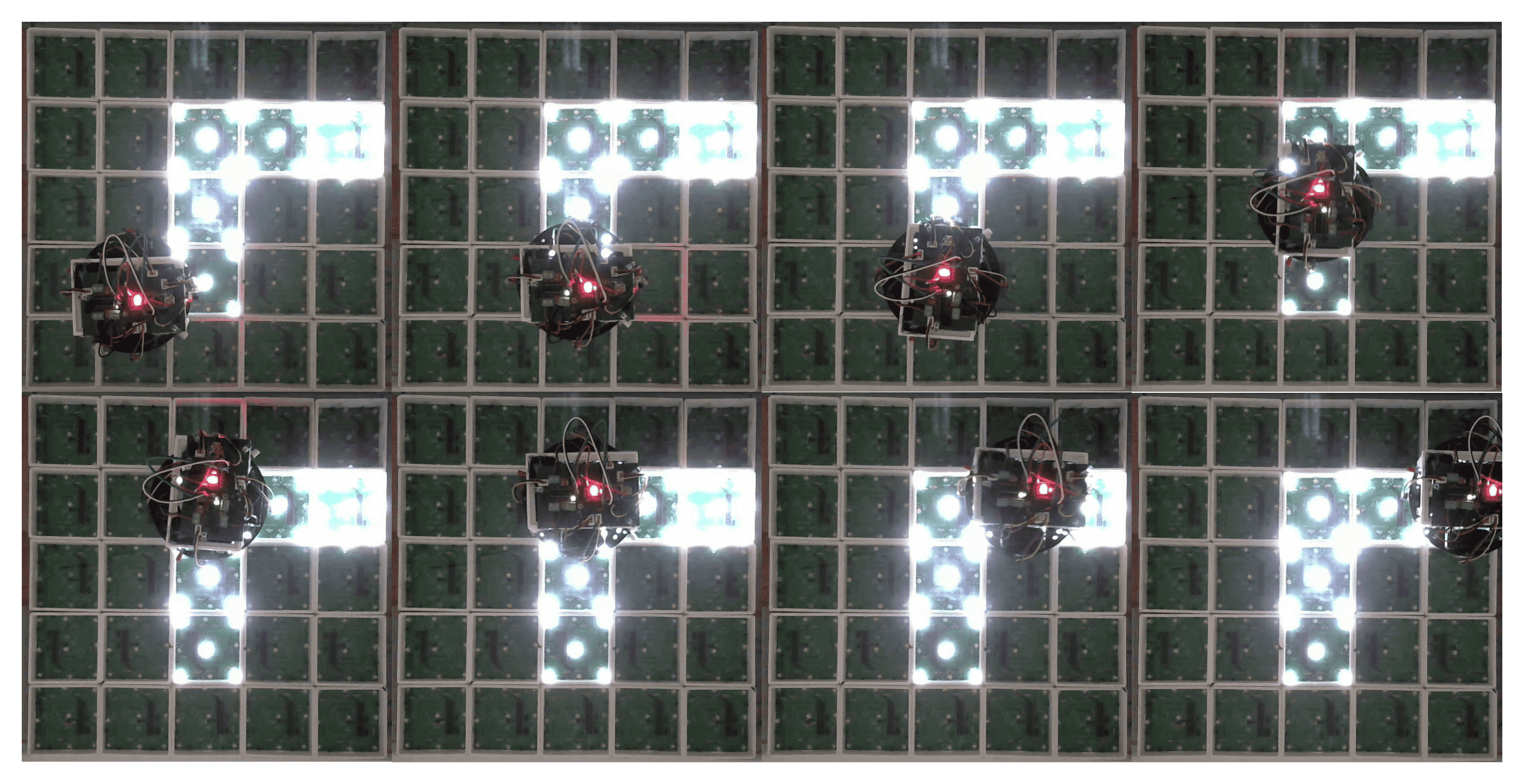}
    \caption{Image showing the sequence of robot following the graded light intensity path, starting from left-top picture to right-bottom picture ~\cite{S1}.}
    \label{fig:multihop2dReal}
\end{figure}

For validating multiple hops along two dimensional (2D) path, the GenGrid
was programmed to provide a constant intensity gradient along the path.
The cells representing the two-dimensional path are configured with light intensities of 0\%, 20\%, 40\%, 60\%, 80\%, and 100\% to achieve a constant intensity difference.
All other cell's lighting was switched off to maintain 0\% intensity, and
hence the robot was expected to follow the programmed 2D path in the platform.
The robot with light sensors positioned on the four sides measures the positive gradient to the light intensity measured at the center of the robot via the attached center light sensor. The robot moves in the direction of the maximum positive gradient.
The robot was positioned at three different starting positions and the experiment
was repeated 20 times, and the probability of the robot reaching the target cell is shown in Figure~\ref{fig:multihop2d}. 
The arrows with color represent
the directional probability of the robot moving in that direction. The time in terms of \emph{t} represents the programmed time for the robot to perform linear and rotational motion to reach the next cell. 
The 3 different initial positions are indicated in Figure~\ref{fig:multihop2d} as~\emph{t = 0} label in the cell. 
The linear movement showcased the highest probability to hop. In contrast, the movement along the sides, which includes an angular rotation of 90\degree depicted a slightly lower probability, which is expected considering the open-loop robot design. The robot movement is driven by a timed PWM signal applied to the two motors. Hence, a mismatch in the motor initiation is likely to introduce orientation error, thereby deviating the path. However, the GenGrid platform offered 2D path configuration capability, and the robot moving in the 2D path was validated with a probability of close to 76\% successfully reaching the target position. One such experimental snapshot is shown in Figure~\ref{fig:multihop2dReal}. A similar experiment was possible with a robot, defining the 2D path in the GenGrid platform is based on the interaction between the robot generating a magnetic field
and GenGrid cells illuminating the path on sensing the same.
The 2D path following robot resembles the pheromone foraging phenomenon, which is successfully demonstrated in the designed novel open-source GenGrid platform.

\begin{figure}[htp]
\centering
    \includegraphics[scale = 0.38]{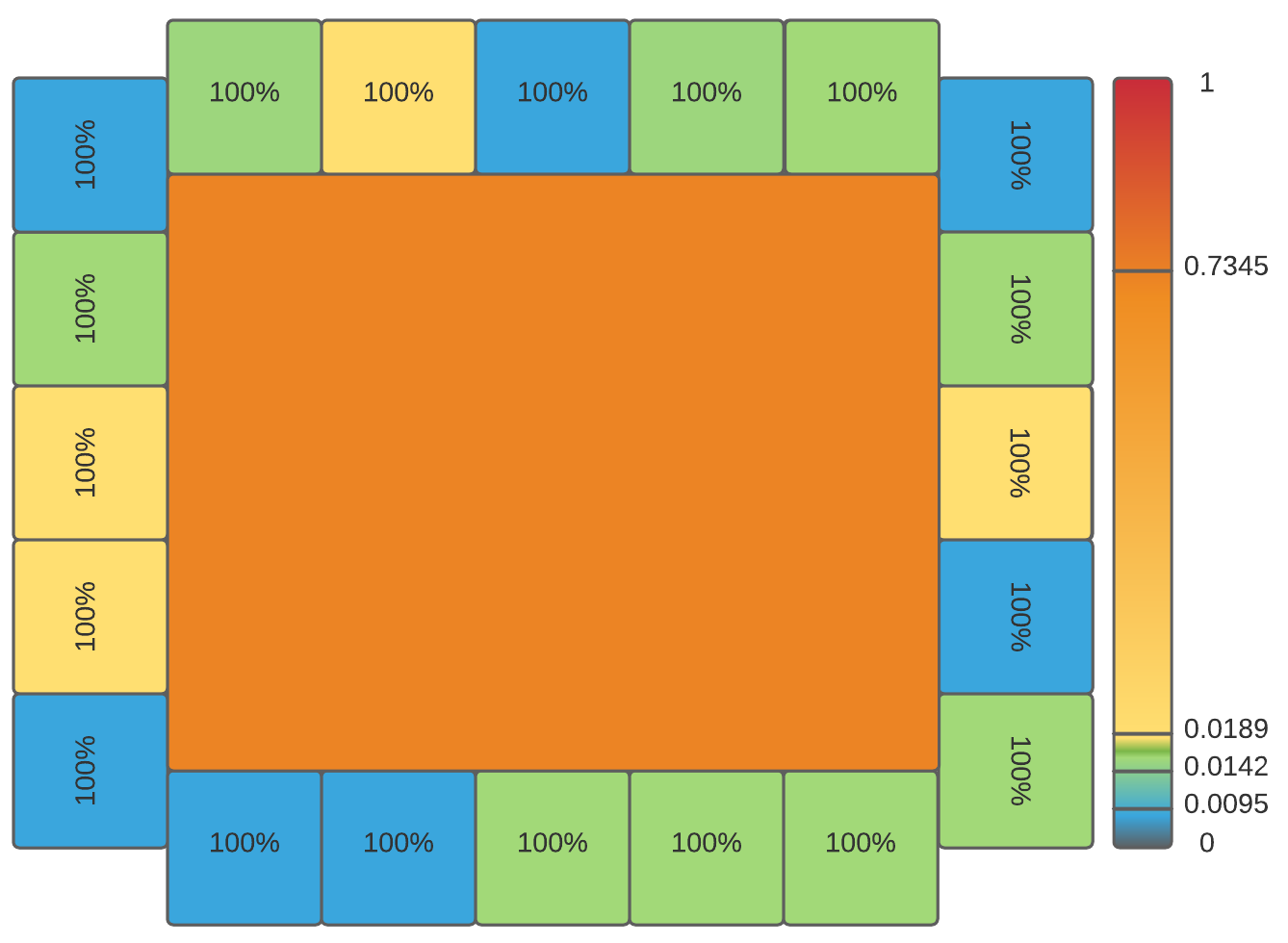}
    \caption{Probability of robot position in a GenGrid map,
    for a wall avoidance experiment ~\cite{S1}.} 
    \label{fig:obstacle}
\end{figure}

In another experiment, the developed GenGrid platform was programmed to provide maximum light intensity around the corners of the grid. The robot was programmed to consider the highly illuminated cells as the virtual wall. The robot was configured to move randomly to the next cell for 5 minutes and avoid moving to the wall-designated cells. The experiment was repeated 20 times by placing the robot in the grid cells away from the corner.
From the 20 experiments, the robot successfully avoided moving to the corner cells 73 mins 27 secs, showing 73\% success. The robot positions staying in the GenGrid platform during the series of experiments is depicted as Red color in Figure~\ref{fig:obstacle}, showcasing the probability of the robot moving to corner cells is very low. 
The light intensity of 100\% is stated in the corner cells indicating that the grid was maximally lit to represent a virtual obstacle for the robot.

\begin{figure}[htp]
    \includegraphics[scale = 0.34]{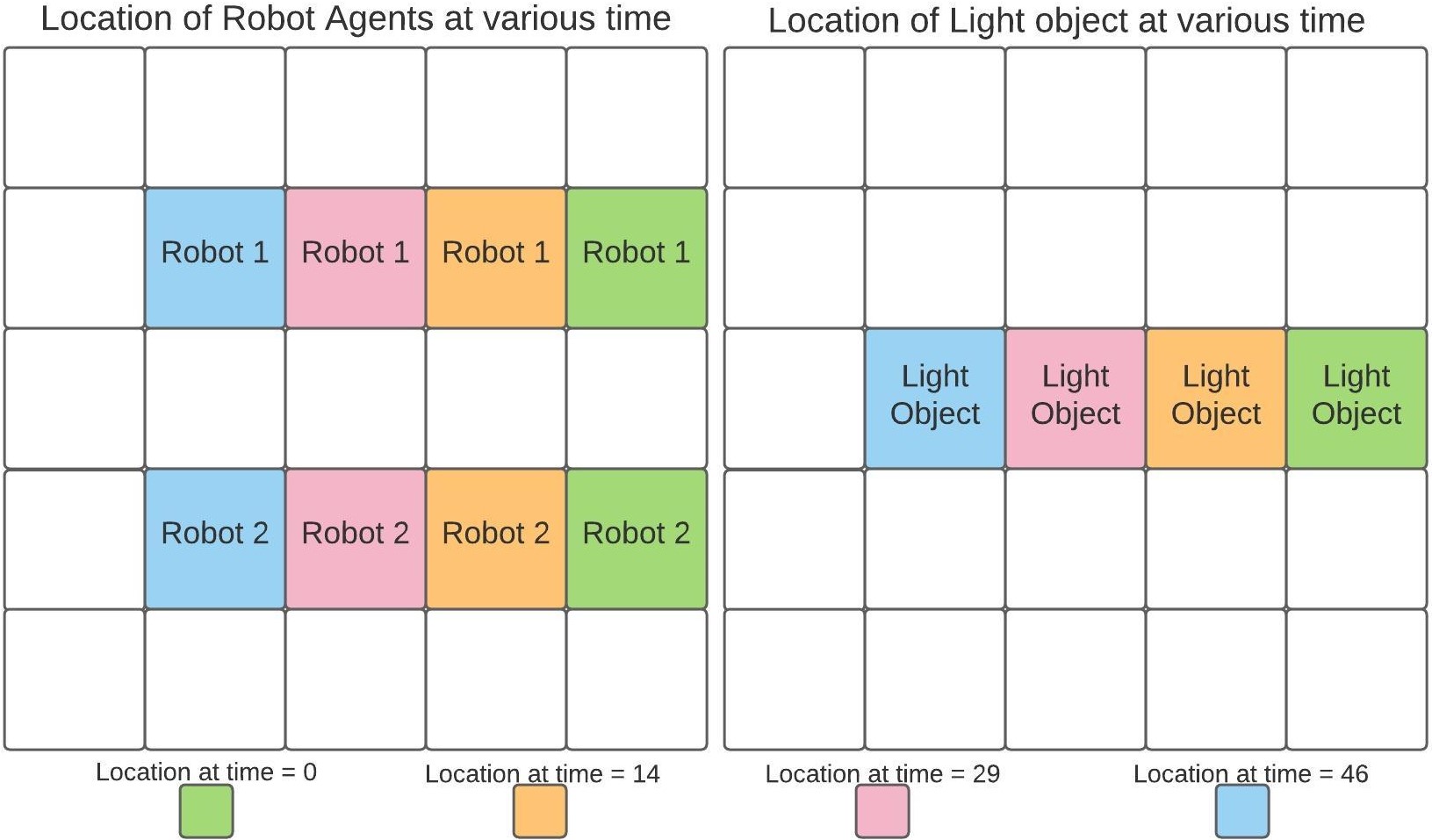}
    \caption{GenGrid map showing the position of two robot's,
    and object movement represented by cell light at various time.~\cite{S1}} 
    \label{fig:collective}
\end{figure}

The GenGrid platform was configured to validate the capability of simulating collective robot transport experiments. The GenGrid was programmed to sense the position of robots on either side of an unoccupied cell using LDR and hall effect sensors and switch on the lights of the empty cell representing the object under transfer. 
The robots were programmed to move in a linear path on detection of the light object, and the action was repeated for four steps to reach the other end of the platform showcasing the collective transport movement.
The movement of the two robots in the GenGrid platform, along with the lighting of the cells along the linear path, is depicted in Figure~\ref{fig:collective} with the color-coded timestamp information, suggesting the possibility of simulating collective transport using the GenGrid platform.

\begin{figure}[htp]
    \includegraphics[scale = 0.28]{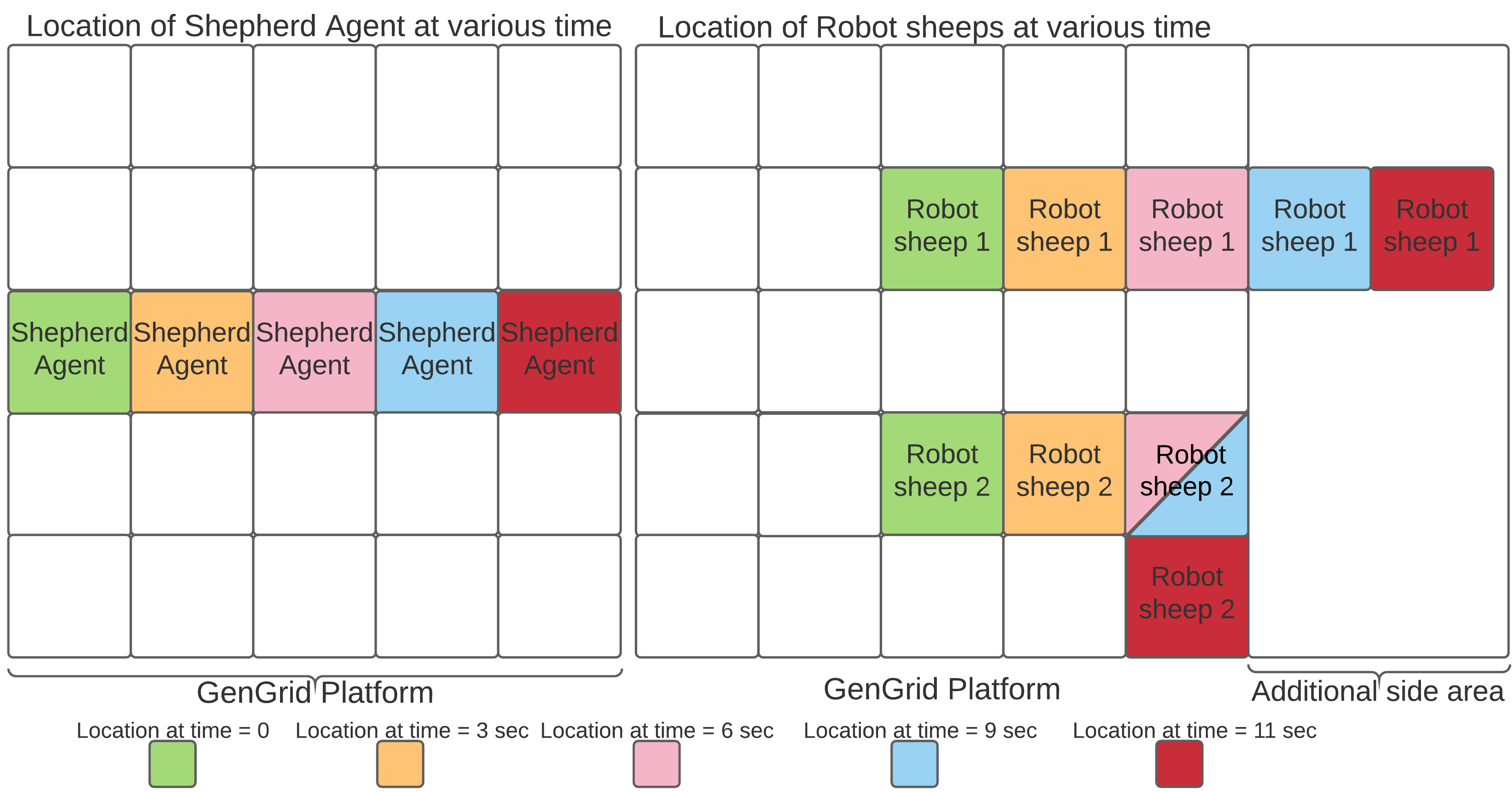}
    \caption{GenGrid map showing the position of two robots and 
    shepherd agent represented by grid light at various time.~\cite{S1}}
    \label{fig:shepherding}
\end{figure}

In another experiment showcasing GenGrid capability in shepherding robots, where an agent interacts and moves the group of robots to the other side, was performed. The activity is commonly observed in sheep, where the caretaker navigates the group of sheep.
The GenGrid is programmed to allow an agent to guide the two robots to another side. A magnet was manually moved in the GenGrid, and the platform recognized the magnetic effect generated, thereby lighting the grid. The robots were programmed to move away from the illuminated grid. Thereby a communication between the agent and robots was established to perform shepherding of robots as shown in Figure~\ref{fig:shepherding}. The timestamp of the agent and two robots are color-coded within the GenGrid cell. The GenGrid platform is suitable for performing shepherding experiments and analyzing the final position of the swarm of robots, which helps the controlled animal navigation application.

\begin{figure}[htp]
    \includegraphics[scale = 0.20]{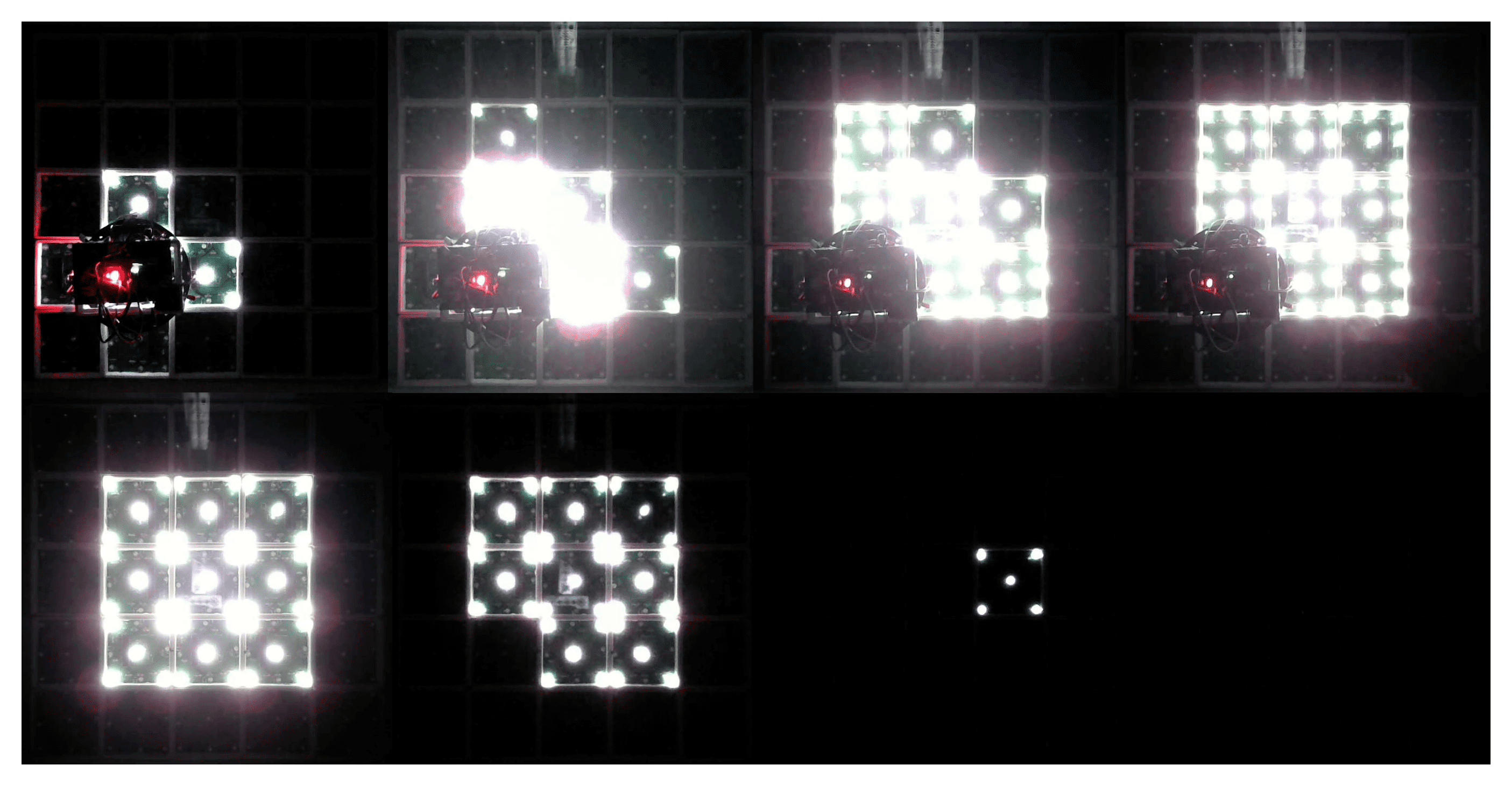}
    \caption{Image showing the sequence of deposition and evaporation of pheromone's in the form of cell lights across the GenGrid platform.~\cite{S1}}
    \label{fig:pheromone}
\end{figure}

The GenGrid's ability to interact with its neighbors is utilized to 
simulate continuous deposition of pheromone released by the robot in the form of light-diffusing to neighborhood cells.
The GenGrid is configured to illuminate all 9 LEDs on detecting a robot in a particular cell. The neighboring cells were programmed to illuminate on detection of light intensity in the neighborhood via LDR sensors or darken otherwise. 
On lifting the robot acting as a pheromone source, the cells tend to diminish with a decrease in light intensity, and eventually, all the cells were unlit.
The continuous pheromone deposition across the neighboring cells and evaporation is depicted in the Figure~\ref{fig:pheromone}, starting from top-left to bottom-right images, showcasing the capability to
perform continuous deposition and evaporation of artificial pheromone experiments using the GenGrid platform.

\section{Conclusions}

A comprehensive open-source platform with minimal sensing and 
messaging capabilities, designed for simulating indoor swarm robotics experiments, were presented.
The designed modular GenGrid platform with minimal resources
of sensing and communication, along with swarm robot design, successfully demonstrated pheromone type swarm capabilities.
The platform is scalable to a larger size in any two dimensional (2D)
directions by attaching the cell to the already configured platform 
without disturbing and affecting other cells in GenGrid. 
The LED light actuation as a feasible communication protocol was employed to interact among the von Neumann neighborhood within GenGrid and send information to other robots in the swarm system.
The magnetic sensor and array of LEDs configured on each cell offered
bidirectional communication between the robotic agent and cell of the platform.
The developed GenGrid platform can offer cue-based communication to leverage the stigmergy phenomenon and enhance the capabilities of swarm robots.
The hardware and software architecture design of the GenGrid system was found optimal considering the realization of different swarm experiments.
The open-source hardware tool is envisioned as a feasible and economical tool for research in multi-robots and swarm robotics.



\bibliographystyle{IEEEtran}
\bibliography{IEEEabrv,paper}

\begin{thebibliography}{10}
\providecommand{\url}[1]{#1}
\csname url@samestyle\endcsname
\providecommand{\newblock}{\relax}
\providecommand{\bibinfo}[2]{#2}
\providecommand{\BIBentrySTDinterwordspacing}{\spaceskip=0pt\relax}
\providecommand{\BIBentryALTinterwordstretchfactor}{4}
\providecommand{\BIBentryALTinterwordspacing}{\spaceskip=\fontdimen2\font plus
\BIBentryALTinterwordstretchfactor\fontdimen3\font minus
  \fontdimen4\font\relax}
\providecommand{\BIBforeignlanguage}[2]{{%
\expandafter\ifx\csname l@#1\endcsname\relax
\typeout{** WARNING: IEEEtran.bst: No hyphenation pattern has been}%
\typeout{** loaded for the language `#1'. Using the pattern for}%
\typeout{** the default language instead.}%
\else
\language=\csname l@#1\endcsname
\fi
#2}}
\providecommand{\BIBdecl}{\relax}
\BIBdecl

\bibitem{colias1}
F.~{Arvin}, J.~C. {Murray}, L.~{Shi}, C.~{Zhang}, and S.~{Yue}, ``Development
  of an autonomous micro robot for swarm robotics,'' in \emph{2014 IEEE
  International Conference on Mechatronics and Automation}, 2014, pp. 635--640.

\bibitem{colias2}
F.~Arvin, S.~Yue, and C.~Xiong, ``Colias-$\phi$: An autonomous micro robot for
  artificial pheromone communication,'' \emph{International Journal of
  Mechanical Engineering and Robotics Research}, vol.~4, no.~4, pp. 349--353,
  2015.

\bibitem{kilobot}
M.~{Rubenstein}, C.~{Ahler}, and R.~{Nagpal}, ``Kilobot: A low cost scalable
  robot system for collective behaviors,'' in \emph{2012 IEEE International
  Conference on Robotics and Automation}, 2012, pp. 3293--3298.

\bibitem{monarobot}
F.~Arvin, J.~Espinosa, B.~Bird, A.~West, S.~Watson, and B.~Lennox, ``Mona: an
  affordable open-source mobile robot for education and research,''
  \emph{Journal of Intelligent \& Robotic Systems}, vol.~94, pp. 761--775,
  2019.

\bibitem{kiloassem}
\BIBentryALTinterwordspacing
M.~Rubenstein, A.~Cornejo, and R.~Nagpal, ``Programmable self-assembly in a
  thousand-robot swarm,'' \emph{Science}, vol. 345, no. 6198, pp. 795--799,
  2014. [Online]. Available:
  \url{https://science.sciencemag.org/content/345/6198/795}
\BIBentrySTDinterwordspacing

\bibitem{swarm1}
M.~Brambilla, E.~Ferrante, M.~Birattari, and M.~Dorigo, ``Swarm robotics: A
  review from the swarm engineering perspective,'' \emph{Swarm Intelligence},
  vol.~7, no.~1, pp. 1--41, 2013.

\bibitem{swarm2}
M.~Dorigo, M.~Birattari, and M.~Brambilla, ``Swarm robotics,''
  \emph{Scholarpedia}, vol.~9, no.~1, p. 1463, 2014.

\bibitem{Hind2013}
I.~Navarro and F.~Matía, ``An introduction to swarm robotics,''
  \emph{International Scholarly Research Notices}, vol. 2013, no. 608164,
  p.~10, 2013.

\bibitem{HeikoBook}
H.~Hamann, \emph{Swarm Robotics: A Formal Approach}, 1st~ed.\hskip 1em plus
  0.5em minus 0.4em\relax Springer International Publishing, 2018.

\bibitem{236}
\BIBentryALTinterwordspacing
K.~Lerman and A.~Galstyan, ``Mathematical model of foraging in a group of
  robots: Effect of interference,'' \emph{Autonomous Robots}, vol.~13, no.~2,
  pp. 127--141, Sep 2002. [Online]. Available:
  \url{https://doi.org/10.1023/A:1019633424543}
\BIBentrySTDinterwordspacing

\bibitem{192}
A.~J. Ijspeert, A.~Martinoli, A.~Billard, and L.~M. Gambardella,
  ``{Collaboration Through the Exploitation of Local Interactions in Autonomous
  Collective Robotics: The Stick Pulling Experiment},'' \emph{Autonomous
  Robots}, vol.~11, no.~2, pp. 149--171, Sep 2001.

\bibitem{239}
A.~F.~T. Winfield, W.~Liu, J.~Nembrini, and A.~Martinoli, ``{Modelling a
  wireless connected swarm of mobile robots},'' \emph{Swarm Intell.}, vol.~2,
  no.~2, pp. 241--266, Dec 2008.

\bibitem{Reina2015}
A.~Reina, R.~Miletitch, M.~Dorigo, and V.~Trianni, ``{A quantitative
  micro{\textendash}macro link for collective decisions: the shortest path
  discovery/selection example},'' \emph{Swarm Intell.}, vol.~9, no.~2, pp.
  75--102, Sep 2015.

\bibitem{218}
J.~Klein, ``{breve: a 3D environment for the simulation of decentralized
  systems and artificial life},'' \emph{Proceedings of the eighth international
  conference on Artificial life}, pp. 329--334, Dec 2002.

\bibitem{142}
B.~P. Gerkey, R.~T. Vaughan, and A.~Howard, ``The player/stage project: Tools
  for multi-robot and distributed sensor systems,'' in \emph{Proceedings of the
  11th International Conference on Advanced Robotics}, 2003, pp. 317--323.

\bibitem{H11}
F.~Mondada, E.~Franzi, and A.~Guignard, ``The development of khepera,''
  \emph{Proceedings of the 1st International Khepera Workshop}, vol. vol. 64 of
  HNI-Verlagsschriftenreihe, pp. 7--14, 1999.

\bibitem{H12}
J.~Pugh, X.~Raemy, C.~Favre, R.~Falconi, and A.~Martinoli, ``{A Fast Onboard
  Relative Positioning Module for Multirobot Systems},'' \emph{IEEE/ASME Trans.
  Mechatron.}, vol.~14, no.~2, pp. 151--162, Feb 2009.

\bibitem{H13}
\BIBentryALTinterwordspacing
F.~Mondada, M.~Bonani, X.~Raemy, J.~Pugh, C.~Cianci, A.~Klaptocz, S.~Magnenat,
  J.-C. Zufferey, D.~Floreano, and A.~Martinoli, ``The e-puck, a robot designed
  for education in engineering,''
  \emph{Proceedings of the 9th Conference on Autonomous Robot Systems and Competitions},
  vol.~1, no.~1, pp. 59--65, 2009. [Online]. Available:
  \url{http://infoscience.epfl.ch/record/135236}
\BIBentrySTDinterwordspacing

\bibitem{H14}
G.~{Caprari} and R.~{Siegwart}, ``Mobile micro-robots ready to use: Alice,'' in
  \emph{2005 IEEE/RSJ International Conference on Intelligent Robots and
  Systems}, 2005, pp. 3295--3300.

\bibitem{H15}
S.~{Kornienko}, O.~{Kornienko}, and P.~{Levi}, ``Minimalistic approach towards
  communication and perception in microrobotic swarms,'' in \emph{2005 IEEE/RSJ
  International Conference on Intelligent Robots and Systems}, 2005, pp.
  2228--2234.

\bibitem{H16}
P.~Valdastri, P.~Corradi, A.~Menciassi, T.~Schmickl, K.~Crailsheim,
  J.~Seyfried, and P.~Dario, ``{Micromanipulation, communication and swarm
  intelligence issues in a swarm microrobotic platform},'' \emph{Rob. Auton.
  Syst.}, vol.~54, no.~10, pp. 789--804, Oct 2006.

\bibitem{H17}
F.~{Mondada}, L.~M. {Gambardella}, D.~{Floreano}, S.~{Nolfi}, J.~. {Deneuborg},
  and M.~{Dorigo}, ``The cooperation of swarm-bots: physical interactions in
  collective robotics,'' \emph{IEEE Robotics Automation Magazine}, vol.~12,
  no.~2, pp. 21--28, 2005.

\bibitem{H18}
A.~Turgut,
  F.~G{\ifmmode\ddot{o}\else\"{o}\fi}k{\ifmmode\mbox{\c{c}}\else\c{c}\fi}e,
  H.~{\ifmmode\mbox{\c{C}}\else\c{C}\fi}elikkanat, L.~Bayindir, and E.~Sahin,
  ``{Kobot: A mobile robot designed specifically for swarm robotics
  research},'' in \emph{Proc. of 7th Int. Conf. on Autonomous Agents and
  Multiagent Systems (AAMAS 2008)}, May 2008, pp. 39--46.

\bibitem{H19}
\BIBentryALTinterwordspacing
J.~D. j.~D. Mclurkin, ``{Stupid robot tricks : a behavior-based distributed
  algorithm library for programming swarms of robots},'' Ph.D. dissertation,
  Massachusetts Institute of Technology, 2004. [Online]. Available:
  \url{https://dspace.mit.edu/handle/1721.1/28550}
\BIBentrySTDinterwordspacing

\bibitem{khaliqstig}
A.~A. {Khaliq} and A.~{Saffiotti}, ``Stigmergy at work: Planning and navigation
  for a service robot on an rfid floor,'' in \emph{2015 IEEE International
  Conference on Robotics and Automation (ICRA)}, 2015, pp. 1085--1092.

\bibitem{cosphi}
F.~{Arvin}, T.~{Krajník}, A.~E. {Turgut}, and S.~{Yue}, ``Cos $\phi$:
  Artificial pheromone system for robotic swarms research,'' in \emph{2015
  IEEE/RSJ International Conference on Intelligent Robots and Systems (IROS)},
  2015, pp. 407--412.

\bibitem{colcosphi}
X.~{Sun}, T.~{Liu}, C.~{Hu}, Q.~{Fu}, and S.~{Yue}, ``Colcos $\phi$: A multiple
  pheromone communication system for swarm robotics and social insects
  research,'' in \emph{2019 IEEE 4th International Conference on Advanced
  Robotics and Mechatronics (ICARM)}, 2019, pp. 59--66.

\bibitem{cos4}
F.~{Fossum}, J.~{Montanier}, and P.~C. {Haddow}, ``Repellent pheromones for
  effective swarm robot search in unknown environments,'' in \emph{2014 IEEE
  Symposium on Swarm Intelligence}, 2014, pp. 1--8.

\bibitem{cos6}
R.~A. {Russell}, ``Ant trails - an example for robots to follow?'' in
  \emph{Proceedings 1999 IEEE International Conference on Robotics and
  Automation (Cat. No.99CH36288C)}, vol.~4, 1999, pp. 2698--2703 vol.4.

\bibitem{cos7}
A.~H. Purnamadjaja and R.~A. Russell, ``{Bi-directional pheromone communication
  between robots},'' \emph{Robotica}, vol.~28, no.~1, pp. 69--79, Jan 2010.

\bibitem{cos8}
R.~A. Russell, ``{Air vortex ring communication between mobile robots},''
  \emph{Rob. Auton. Syst.}, vol.~59, no.~2, pp. 65--73, Feb 2011.

\bibitem{cos9}
R.~Fujisawa, S.~Dobata, K.~Sugawara, and F.~Matsuno, ``{Designing pheromone
  communication in swarm robotics: Group foraging behavior mediated by chemical
  substance},'' \emph{Swarm Intell.}, vol.~8, no.~3, pp. 227--246, Sep 2014.

\bibitem{cos10}
Herianto, T.~Sakakibara, and D.~Kurabayashi, ``{Artificial pheromone system
  using RFID for navigation of autonomous robots},'' \emph{J. Bionic Eng.},
  vol.~4, no.~4, pp. 245--253, Dec 2007.

\bibitem{cos12}
F.~Arvin, A.~E. Turgut, F.~Bazyari, K.~B. Arikan, N.~Bellotto, and S.~Yue,
  ``{Cue-based aggregation with a mobile robot swarm: a novel fuzzy-based
  method},'' \emph{Adaptive Behavior}, vol.~22, no.~3, pp. 189--206, May 2014.

\bibitem{cos13}
O.~{Holland} and C.~{Melhuish}, ``An interactive method for controlling group
  size in multiple mobile robot systems,'' in \emph{1997 8th International
  Conference on Advanced Robotics. Proceedings. ICAR'97}, 1997, pp. 201--206.

\bibitem{cos14}
S.~Garnier, M.~Combe, C.~Jost, and G.~Theraulaz, ``{Do Ants Need to Estimate
  the Geometrical Properties of Trail Bifurcations to Find an Efficient Route?
  A Swarm Robotics Test Bed},'' \emph{PLoS Comput. Biol.}, vol.~9, no.~3, p.
  e1002903, Mar 2013.

\bibitem{cos15}
F.~Arvin, K.~Samsudin, A.~R. Ramli, and M.~Bekravi, ``{Imitation of Honeybee
  Aggregation with Collective Behavior of Swarm Robots},'' \emph{International
  Journal of Computational Intelligence Systems}, vol.~4, no.~4, pp. 739--748,
  Jun 2011.

\bibitem{cos16}
T.~Schmickl, R.~Thenius, C.~Moeslinger, G.~Radspieler, S.~Kernbach,
  M.~Szymanski, and K.~Crailsheim, ``{Get in touch: cooperative decision making
  based on robot-to-robot collisions},'' \emph{Auton. Agent. Multi-Agent
  Syst.}, vol.~18, no.~1, pp. 133--155, Feb 2009.

\bibitem{cos17}
K.~{Sugawara}, T.~{Kazama}, and T.~{Watanabe}, ``Foraging behavior of
  interacting robots with virtual pheromone,'' in \emph{2004 IEEE/RSJ
  International Conference on Intelligent Robots and Systems (IROS) (IEEE Cat.
  No.04CH37566)}, vol.~3, 2004, pp. 3074--3079 vol.3.

\bibitem{cos18}
S.~{Garnier}, F.~{Tache}, M.~{Combe}, A.~{Grimal}, and G.~{Theraulaz}, ``Alice
  in pheromone land: An experimental setup for the study of ant-like robots,''
  in \emph{2007 IEEE Swarm Intelligence Symposium}, 2007, pp. 37--44.

\bibitem{kilogridjournal}
G.~Valentini, A.~Antoun, M.~Trabattoni, B.~Wiandt, Y.~Tamuraand, E.~Hocquard,
  V.~Trianni, and M.~Dorigo, ``Kilogrid: a novel experimental environment for
  the kilobot robot,'' \emph{Swarm Intelligence}, vol. 12.3, pp. 245--266,
  2018.

\bibitem{S1}
\BIBentryALTinterwordspacing
P.~Kedia and M.~Rao, ``{Supplementary Material: Experimental Videos and
  Schematic files of GenGrid}.'' [Online]. Available:
  \url{https://praked.github.io/portfolio/portfolio-1/}
\BIBentrySTDinterwordspacing

\end{thebibliography}

\end{document}